%% file: nips_2018_prob_verify.tex
\documentclass{article}

 \usepackage[final]{nips_2018}

\usepackage{hyperref}
\usepackage{url}
\usepackage{enumitem}
\usepackage{amsmath}
\usepackage{amsfonts}
\usepackage{amssymb}
\usepackage{mathtools}
\usepackage{graphicx}
\usepackage{subfigure}	
\usepackage{mathrsfs}
\usepackage{amsthm}
\usepackage{algorithmic}
\usepackage{natbib}
\usepackage{algorithm}
\usepackage{bm}
\usepackage{hyperref}
\usepackage{verbatim}
\usepackage{wrapfig}
\newtheorem{theorem}{Theorem}

\usepackage{listings}
\usepackage{color}
\usepackage[utf8]{inputenc} % allow utf-8 input
\usepackage[T1]{fontenc}    % use 8-bit T1 fonts
\usepackage{hyperref}       % hyperlinks
\usepackage{url}            % simple URL typesetting
\usepackage{booktabs}       % professional-quality tables
\usepackage{amsfonts}       % blackboard math symbols
\usepackage{nicefrac}       % compact symbols for 1/2, etc.
\usepackage{microtype}      % microtypography

\title{Verification of deep probabilistic models}

\author{
  Krishnamurthy (Dj) Dvijotham \\
  DeepMind\\
  London, UK \\
  \texttt{dvij@cs.washington.edu} \\
  \And
  Marta Garnelo \\
  DeepMind\\
  London, UK \\
  \texttt{garnelo@google.com} \\
  \And
  Alhussein Fawzi \\
  DeepMind\\
  London, UK \\
  \texttt{afawzi@google.com} \\
  \And
  Pushmeet Kohli \\
  DeepMind\\
  London, UK \\
  \texttt{pushmeet@google.com}
}

\input{notations.tex}

\begin{document}

\maketitle

\begin{abstract}
Probabilistic models are a critical part of the modern deep learning toolbox - ranging from generative models (VAEs, GANs), sequence to sequence models used in machine translation and speech processing to models over functional spaces (conditional neural processes, neural processes). Given the size and complexity of these models, safely deploying them in applications requires the development of tools to analyze their behavior rigorously and provide some guarantees that these models are consistent with a list of desirable properties or \emph{specifications}. For example, a machine translation model should produce semantically equivalent outputs for innocuous changes in the input to the model. A functional regression model that is learning a distribution over monotonic functions should predict a larger value at a larger input. Verification of these properties requires a new framework that goes beyond notions of verification studied in deterministic feedforward networks, since requiring worst-case guarantees in probabilistic models is likely to produce conservative or vacuous results. We propose a novel formulation of verification for deep probabilistic models that take in conditioning inputs and sample latent variables in the course of producing an output: We require that the output of the model satisfies a linear constraint \emph{with high probability over the sampling of latent variables and for every choice of conditioning input to the model}. We show that rigorous lower bounds on the probability that the constraint is satisfied can be obtained efficiently. Experiments with neural processes show that several properties of interest while modeling functional spaces can be modeled within this framework (monotonicity, convexity) and verified efficiently using our algorithms.
\end{abstract}

\section{Introduction}
Several deep learning models are inherently probabilistic in nature. Common examples include generative models like variational autoencoders \citep{kingma2013auto, sohn2015learning}, functional regression models like neural processes \citep{garnelo2018conditional, garnelo2018neural}, models with attention mechanisms\citep{xu2015show} and sequence to sequence models \citep{van2016wavenet}. Given the widespread application of such models, it is of interest to develop tools that formally verify consistency of these models with specifications of interest (for example, a functional regression model should only produce functions within the class of functions of interest - monotonic, convex etc., or an attention model only attends to a relevant parts of the input). However, verification in a worst-case sense over inputs and probabilistic components is likely to lead to meaningless or trivial results. For example, models that have Gaussian latent variables can produce garbage outputs if the latent variables take extreme large positive or negative values (which is possible, albeit unlikely). Thus, a different form of verification is necessary in such models. We propose the following notion here - for all conditioning inputs to the model within a given set of interest, with high probability over the sampling of latent variables, the output of the network satisfies a certain property. Formally, the contributions of this paper are:
\begin{itemize}
    \item Formalizing a notion of verification that is applicable to deep probabilistic models and extending verification algorithms to apply to deep probabilistic models, like VAEs \citep{kingma2013auto}, conditional VAEs \citep{sohn2015learning}, conditional neural processes \citep{garnelo2018conditional} and neural processes \citep{garnelo2018neural}.
    \item Experimental results on neural processes \citep{garnelo2018neural} showing that the verification algorithm can prove interesting properties about a learnt  neural process model.
\end{itemize}

\input{technical_neuralnet.tex}

\input{experiments.tex}

\section{Conclusions}
We presented a novel verification formulation and algorithm for verification of deep probabilisitc models. In future work, we will study more complex models (CVAEs, GANs), specifications (for example, verifying disentangled representations) and integration of verification into training .

\bibliography{references}
\bibliographystyle{plainnat}

\end{document}

%% file: notations.tex
\newcommand{\br}[1]{\left({#1}\right)}

\newcommand{\lambdall}{{\boldsymbol{\lambda}}}

\newcommand{\R}{\mathbb{R}}

\newcommand{\norm}[1]{\left\|{#1}\right\|}

%% file: technical_neuralnet.tex
\section{Verification of deep probabilistic models}
We formulate the problem of verification of deep probabilistic models. We focus on a special architecture here (extension to more general models will be pursued in future work). Since verification is concerned with checking properties of ML models after they have been trained, we are primarily interested in the decoder of deep probabilistic models like conditional VAEs \citep{sohn2015learning} and (conditional) neural processes \citep{garnelo2018neural, garnelo2018conditional}. A typical architecture for the decoder phase of these models is a neural network that receives as input both a conditioning input, denoted $x$, and a latent variable sampled from a distribution generated by the encoder, denoted $z$. In most architectures, it is common for the latent variable $z$ to be sampled from a Gaussian distribution. 

Thus, the output of the model is given by $f\br{x, z}, z \sim \mathcal{N}\br{0, I}$ where $f$ is a decoder network.\footnote{Note here that the assumption $z$ is sampled from a standard normal distribution is not restrictive, since if $z \sim \mathcal{N}\br{\mu, \Sigma}$ we can always reparameterize $z=\Sigma^{1/2}\tilde{z}+ \mu$ where $\tilde{z} \sim \mathcal{N}\br{0, I}$. This framework can also extend to models with stochasticity entering at intermediate layers and to models with multiplicative interactions - such extensions will be pursued in an extended version of this paper.}. We will assume that $f$ is composed of layers of transformations of the form $x_{k+1}=W_k h_k(x_k) + b_k$ where $h_k$ is a component-wise nonlinearity (like ReLU/sigmoid/tanh) and $(W_k, b_k)$ define a linear transformation. Note that this can capture convolutional networks as well, since a convolution is simply a specially structured linear transformation. We will use $x$ to denote either single inputs or a stacked vector of several inputs. If $x$ is a stacked vector, we assume that the network outputs a stacked vector of outputs corresponding to each input in $x$.

\paragraph{Verification problem:}  The verification problem, parameterized by a vector of coefficients $c$ and scalar $d, \epsilon$, is defined as follows: Check that the predictor (decoder) function $f$ satisfies the following:
\begin{align}
\mathbb{P}\br{c^T f\br{x, z} + d \geq 0} \leq \epsilon \quad \forall x \in \mathcal{X} \label{eq:prob_verify}
\end{align}
where $\mathcal{X}$ is a set of inputs of interest and $\mathbb{P}$ refers to the probability measure induced by the random variable $z$. Thus, we require that the output $y$ of the network satisfies the constraint $c^T y + d \leq 0$ with probability at least $1-\epsilon$ for each input $x \in \mathcal{X}$.

In this paper, we will primarily work with neural processes that perform functional regression, where $f$ predicts the value of an unknown function at a target point given a set of context points and corresponding function values. In this setting, it is of interest to verify that the outputs of the model satisfy properties of interest that are known to be satisfied for the class of functions being modeled:\\
\emph{Boundedness:} In several domains, the functions of interest ought to be bounded above (or below) by a fixed number $a$ - for example, if the class of functions are cumulative distribution functions, they must be bounded between $0$ and $1$. This can be modeled as:
\[\mathbb{P}\br{f\br{x, z} - a \geq 0} \leq \epsilon \quad \forall x \in \mathcal{X}\]
or, $f\br{x, z} \leq a$ with high probability. \\
\emph{Monotonicity:} Monotonicity is another property of interest (that has to be true for CDFs):
\[\mathbb{P}\br{f\br{x_1, z} - f\br{x_2, z} \geq 0} \leq \epsilon \quad \forall (x_1, x_2) \in \mathcal{X}, x_1 \leq x_2\]
or that $f\br{x_1, z} \leq f\br{x_2, z}$ with high probability. Denoting $x=(x_1, x_2)$ (the stacked vector of inputs) and choosing $c=(1, -1)$, this can again be modeled in our framework. \\
\emph{Midpoint-Convexity:} 
\[\mathbb{P}\br{\frac{f\br{x_1, z} + f\br{x_2, z}}{2} - f\br{\frac{x_1 + x_2}{2}, z}\geq 0} \leq \epsilon \quad \forall (x_1, x_2) \in \mathcal{X}\]
or that $\frac{f\br{x_1, z} + f\br{x_2, z}}{2} \leq f\br{\frac{x_1 + x_2}{2}}$ with high probability (midpoint convexity is equivalent to convexity for continuous and bounded functions \citep{boyd2004convex}). Denoting $x=(x_1, x_2, \frac{x_1+x_2}{2})$ (the stacked vector of inputs) and choosing $c=\left(\frac{1}{2}, \frac{1}{2}, -1\right)$, this can again be modeled in our framework. This can be particularly useful in the context of Bayesian optimization \citep{frazier2018tutorial}, where the posterior over functions produced by a model should be easy to optimize to find the next target point to acquire new information.

\section{Verification algorithm}

Ensuring that \eqref{eq:prob_verify} is satisfied is challenging for two reasons: The space of inputs $\mathcal{X}$ can be large and checking that the constraint holds is challenging (it has been shown to be NP-hard to check even approximately in \citep{weng2018towards}), in the worst case requiring a brute force enumeration approach using SMT \citep{katz2017reluplex} or mixed integer programming solvers \citep{bunel2017piecewise}, which have not yet been scaled to realistically sized modern deep learning systems. However, for probabilistic models, an additional source of difficulty is that computing the probability $\mathbb{P}\br{c^T f(x, z) + d \geq 0}$, even for a fixed $x$, involves solving an intractable probabilistic inference problem. While this can be estimated via Monte-Carlo sampling, since in the context of verification we are interested in rigorous guarantees, estimates do not suffice. We develop an approach to overcome these challenges.

%We resolve first problem can be solved by constructing upper bounds on the optimal value of the problem $\max_{x \in \mathcal{X}} c^T f(x, z) + d$ for a given value of $z$ - this can be done using the approach developed in \citep{DjDual}, which provides an upper bound $g\br{\lambda, z}$ dependent on a choice of \emph{dual variables} $\lambda$. The second problem we resolve by treating treating $g\br{\lambda, z}$ as a function of $z$ for any fixed value of $\lambda$ - due to the linear dependence of $g\br{\lambda, z}$ on $z$ (which follows from the how the dual function $g$ is constructed), the probability $\mathbb{P}\br{g\br{\lambda, z} \geq 0}$ can be computed analytically. Since $g$ is an upper bound on $f$, this probability is an upper bound on the probability \eqref{eq:prob_verify}, and hence if $\mathbb{P}\br{g\br{\lambda, z} \geq 0} \leq \epsilon$ then \eqref{eq:prob_verify} is true. Finally, to get the best possible upper bound, we simply optimize $\mathbb{P}\br{g\br{\lambda, z} \geq 0}$ wrt $\lambda$ - if the optimal value is smaller than $\epsilon$, the verification has successfully certified \eqref{eq:prob_verify}. 

Consider the optimization problem $\max_{x \in \mathcal{X}} c^T f(x, z) + d$. Exploiting the fact that $f$ is composed of multiple layers of simple transformations, we can write this as
\begin{align*}
\max_{x_0, x_1, \ldots, x_K} & c^T x_K + d \\
\text{Subject to } \ & x_{k+1} = W_k h_k(x_k) + b_k, \quad k=1, \ldots, K-1 \\
& x_1 = W_0 x_0 + \tilde{W_0}z + b_0, x_0 \in \mathcal{X}
\end{align*}
where $(W_0, \tilde{W_0}, b_0)$ define the first linear layer that takes both the conditioning input $x$ and the latent variable $z$.

We first condition on the event $z \in [\alpha, \beta]$ and assume that $\mathcal{X}=[l_0, u_0]$ is a set defined by interval constraints on the input $x$. Given these, we can infer bounds for all the intermediate layers using the techniques described in \citep{DjDual, weng2018towards} to obtain $l_k \leq x_k \leq u_k$\footnote{Note that $l_k, u_k$ depend on $\alpha, \beta$.} Since $x_K=W_{K-1} h_{K-1}\br{x_{K-1}}+b_{K-1}$, we can further simplify the problem to
\begin{align*}
\max_{x_0, x_1, \ldots, x_{K-1}} & c^T\br{W_{K-1} h_{K-1}\br{x_{K-1}}+b_{K-1}} + d \\
\text{Subject to }\ & x_{k+1} = W_k h_k(x_k) + b_k, \quad k=1, \ldots, K-2 \\
& x_1 = W_0 x_0 + \tilde{W_0}z+b_0 \\
& l_k \leq x_k \leq u_k, \quad k=0, \ldots, K-1 \\
& \alpha \leq z \leq \beta
\end{align*}
Taking the Lagrangian dual of this problem and rearranging terms using a construction similar to \citet{DjDual}, we obtain
\begin{align}
G\br{\lambdall, z} = d + \nu_0^Tz + \sum_{k=0}^{K-1} \max_{l_k \leq x_k \leq u_k} \nu_k^T h_k\br{x_k} - \lambda_{k-1}^T x_k  + \lambda_k^T b_k   \label{eq:G}
\end{align}
where $\nu_k=W_k^T \lambda_k$ and $\lambda_{K-1}=c, \lambda_{-1}=0$ and $\lambdall=\{\lambda_k\}_{k=0}^{K-2} \in \R^p$ ($p$ is the sum of sizes of the intermediate layers). While the bounds on $z$ do not appear explicitly, $G$ depends on them via $l_k, u_k$

Since the $h_k$ are component-wise nonlinearities, the maximization can be solved independently for each dimension of $x_k$ (typically in closed-form for most common activation functions, as described in \citep{DjDual}) and thus the dual function can be computed easily. By weak duality \citep{boyd2004convex}, we have $ 
c^T f\br{x, z} +d \leq G\br{\lambdall, z} \quad \forall x \in \mathcal{X}, \lambdall \in \R^p, z\in [\alpha, \beta]$. Thus,
\begin{align*}
    \mathbb{P}\br{c^T f\br{x, z} +d \geq 0 | z \in [\alpha, \beta]} \leq \mathbb{P}\br{G\br{\lambdall, z} \geq 0 | z \in [\alpha, \beta]}
\end{align*}
Denote the events $c^T f\br{x, z} +d \geq 0$, $z \in [\alpha, \beta]$ and $G\br{\lambdall, z} \geq 0$ as $A, B, C$ respectively. We then have
\begin{align*}
    & P(A) = P(A|B)P(B) + P(A|\neg B) P(\neg B)  \leq P(C|B)P(B) + P(\neg B)  \leq P(C) + P(\neg B)
\end{align*}
Thus, $\mathbb{P}\br{c^T f\br{x, z} +d \geq 0} \leq \mathbb{P}\br{G\br{\lambdall, z} \geq 0} + \mathbb{P}\br{z \not\in [\alpha, \beta]}$. Define $g\br{\lambdall} = G\br{\lambdall, z} - \nu_0^T z$ so that
\begin{align*}
    & \mathbb{P}\br{G\br{\lambdall, z} \geq 0} = \mathbb{P}\br{\nu_0^T z \geq -g\br{\lambdall}} = \mathbb{P}\br{\omega \geq -\frac{g\br{\lambdall}}{\norm{\nu_0}}} = \frac{1}{2}\mathrm{erfc}\br{-\frac{g\br{\lambdall}}{\sqrt{2}\norm{\nu_0}}}
\end{align*}
where $\omega$ is a standard normal random variable and $\mathrm{erfc}$ is the Gaussian complementary error function.
\begin{theorem}
The probability of violating the specification \eqref{eq:prob_verify} can be bounded as:
\begin{align}
\mathbb{P}\br{c^T f\br{x, z} +d \geq 0 } \leq \frac{1}{2}\mathrm{erfc}\br{-\frac{g\br{\lambdall}}{\sqrt{2}\norm{\nu_0}}} + \mathbb{P}\br{z \not\in [\alpha, \beta]} \label{eq:prob_bound}
\end{align} 
for any choice of $\lambdall, \alpha, \beta$ with $\alpha \leq \beta$.
\end{theorem}
We note that since the RHS of \eqref{eq:prob_bound} is differentiable wrt $\lambdall, \alpha, \beta$, it can be optimized using gradient descent to find the tightest possible bound. By parameterizing $\beta=\alpha + \eta^2$, this can be done by solving an unconstrained optimization problem over $\alpha, \eta, \lambdall$ using gradient descent. Although nonconvex, in practice we find that this optimization can be solved efficiently.

%% file: experiments.tex
\section{Experiments}
A neural process\citep{garnelo2018neural} may be viewed as a neural approximation of a gaussian process that learns a posterior distribution over functions. The NP is trained on functions that are cumulative distribution functions (CDFs) of beta distributions with varying parameters - at the end of training, we expect that the neural process has learned a posterior distribution that, with high probability, produces samples that look like a CDF. The decoder we use is a fully connected network with a 3 hidden layer of $64$ units each and relu activations.

\begin{wrapfigure}{l}{.4\textwidth}
\vspace{-20pt}
    \includegraphics[width=.4\textwidth]{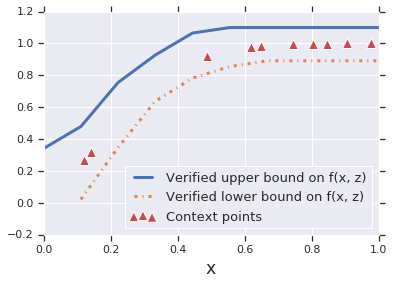} \vspace{-20pt}
    \caption{Verification of NP predictions}
    \label{fig:upper_bounds}
\end{wrapfigure}
At test time, the neural process is given a new set of context points (pairs of inputs and function values) form a previously unseen CDF and asked to predict the value of the CDF at a set of target points. The context points are plotted as red triangles in figure \ref{fig:upper_bounds}. Our verification is on the prediction of the NP at the unseen target points. We use the verification algorithm to compute bounds on the probability that when the target input is in a given range, the predicted output is above or below a certain bound.  

We choose a threshold of $\epsilon=.01$ and find the smallest bound $a$ such that $f\br{x, z} \leq a$ with probability at least $1-\epsilon$ as a function of a range specified on $x$. Specifically, we consider intervals of width $.02$, ie $x\in \mathcal{X}=[\delta, \delta + .02]$ and plot the value of $a$ as a function of $\delta$. Similarly, we find the largest lower bound $b$ such that $f\br{x, z} \geq b$ with probability at least $1-\epsilon$. The upper bound is plotted as a solid blue line, and the lower bound as a dashed yellow line. The results show that the values of $a, b$ increase with $\delta$, and generally conform with the properties of a CDF, thus showing that our verification algorithm is indeed able to prove properties we expect to be true of the model.